\documentclass{article}

\usepackage{microtype}
\usepackage{graphicx}
\usepackage{subfigure}
\usepackage{booktabs} 

\usepackage[dvipsnames]{xcolor}
\usepackage{pbox}
\usepackage{array,multirow}

\usepackage{hyperref}

\usepackage[accepted]{icml2021}

\icmltitlerunning{Multimodal Conditionality for Natural Language Generation}

\begin{document}

\twocolumn[
\icmltitle{Multimodal Conditionality for Natural Language Generation}



\icmlsetsymbol{equal}{*}

\begin{icmlauthorlist}
\icmlauthor{Aashish Jain}{equal,sfe}
\icmlauthor{Michael Sollami}{equal,sfe}
\end{icmlauthorlist}

\icmlaffiliation{sfe}{Salesforce Einstein}

\icmlcorrespondingauthor{Michael Sollami}{msollami@salesforce.com}
\icmlcorrespondingauthor{Aashish Jain}{aashishjain@salesforce.com}

\icmlkeywords{Machine Learning, ICML}

\vskip 0.3in
]



\printAffiliationsAndNotice{\icmlEqualContribution} 

\begin{abstract}


Large scale pretrained language models have demonstrated state-of-the-art performance in language understanding tasks. Their application has recently expanded into multimodality learning, leading to improved representations combining vision and language. However, progress in adapting language models towards conditional Natural Language Generation (NLG) has been limited to a single modality, generally text. We propose MAnTiS, Multimodal Adaptation for Text Synthesis, a general approach for multimodal conditionality in transformer-based NLG models. In this method, we pass inputs from each modality through modality-specific encoders, project to textual token space, and finally join to form a conditionality prefix. We fine-tune the pretrained language model and encoders with the conditionality prefix guiding the generation. We apply MAnTiS to the task of product description generation, conditioning a network on both product images and titles to generate descriptive text. We demonstrate that MAnTiS outperforms strong baseline approaches on standard NLG scoring metrics. Furthermore, qualitative assessments demonstrate that MAnTiS can generate human quality descriptions consistent with given multimodal inputs.

\end{abstract}

\begin{figure*}[ht]
\vskip 0.2in
\begin{center}
\centerline{\includegraphics[trim=200 55 0 37,
scale=0.62]{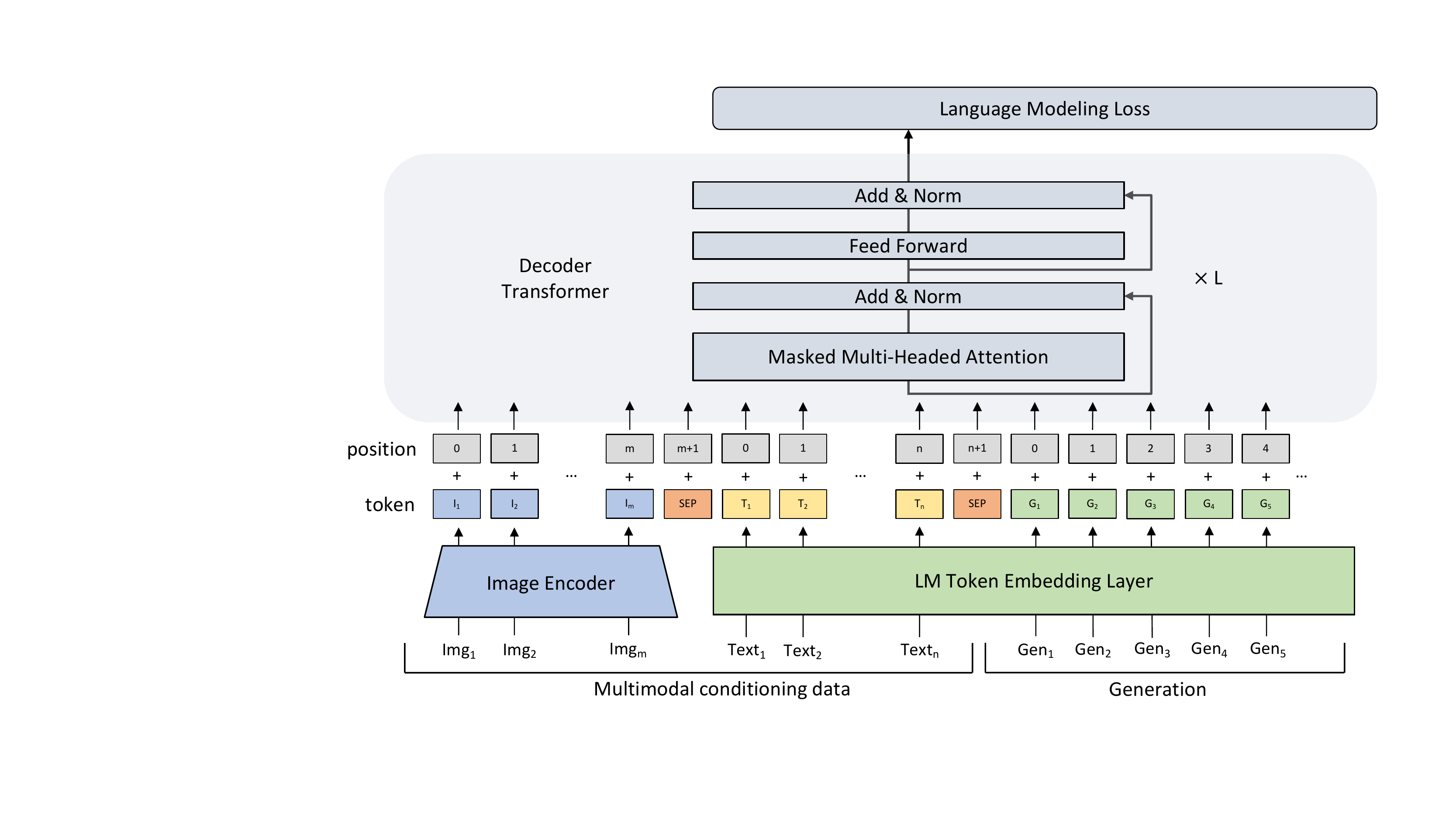}}
\caption{An overview of the MAnTiS architecture. Conditioning images are passed as input through an image encoder and mapped to textual token space of language model. Input text is encoded using the language model's encoder and together with image tokens form the conditionality prefix. The language modeling loss is computed only for the text tokens. Here $m$ and $n$ represent the number of input images and text tokens respectively and $L$ is the number of decoder transformer layers.}
\label{fig1}
\end{center}
\vskip -0.2in
\end{figure*}

\section{Introduction}
\label{introduction}

The use of transfer learning techniques in Natural Language Processing (NLP) significantly improves previous state of the art methods across a wide range of NLP tasks \cite{dai2015semi, devlin2018bert, howard2018universal, radford2019language, brown2020language}. In this setting a transformer-based language model is pretrained on large unlabelled corpra and then fine-tuned on supervised data together with a task-related head \cite{devlin2018bert}. Such approaches are prominent in Natural Language Understanding (NLU) tasks, but remain less explored for text generation.

Transfer learning methods have recently been applied to the joint learning of multiple modalities, where both image and text based inputs are pretrained together \cite{lu2019vilbert, li2020unicoder, su2019vl, chen2020uniter, li2019visualbert}. In these approaches, learning combined representations of visual and textual data during pretraining instead of task specific training, leads to better semantic representations. Due to state-of-the-art performance and straightforward downstream training, it is fast becoming the default method for multimodal tasks like visual question answering, visual entailment, and caption-based image retrieval.

A natural extension to this approach would adapt pretrained language models for conditional Natural Language Generation (NLG) with multimodal conditionality. This can be achieved in an encoder-decoder framework where the encoder learns to embed conditionality while the pretrained decoder would modify the generation based on this encoding. Earlier work suggests this works well for tasks where generation depends on purely textual information \cite{golovanov2019large, zhang2019pretraining, song2019mass}. Recent work used other modalities like image or class information \citet{ziegler2019encoder} to guide the generation of pretrained models. However, that work considered only a single modality and required the introduction of new parameters within the pretrained model that could adversely affect generation capability.

In this work, we propose MAnTiS, a general approach to adapt transformer-based language models into multimodal conditional NLG models. We encode each modality type using specific encoders, joined to form a conditionality prefix, with separator tokens delimiting each modality. During fine-tuning the decoder uses the prefix as history and predicts outputs in a continuous fashion. Because the prefix is decoder independent, the generation can be conditionalized towards any modality. We drew inspiration from \citet{kiela2019supervised} which shows that self-supervised unimodal transformer models are capable of learning context between different modalities through supervised learning for classification tasks.\\

We demonstrate the effectiveness of this approach on a fashion captioning dataset \cite{yang2020fashion}, where given a product's name and image, the model generates an e-commerce relevant description. We compare generations against competing approaches that rely on injecting conditionality vectors into pretrained language models. Through this, we found that MAnTiS outperforms other models. We perform both quantitative and qualitative experiments and show the effectiveness of this approach without requiring any complex model stitching. Extension of MAnTiS to any modality type  is straightforward to implement for any transformer-based pretrained model. We thus provide a strong baseline approach for future transfer learning in NLG.

\begin{table*}
  \caption{A sample entry from the FACAD dataset of fashion products (metadata not shown). }
  \label{table1}
  \vskip 0.15in
  \centering
  \begin{tabular}{p{3cm}|p{7.5cm}|p{5cm}}
    \toprule
    \textbf{Product Title} & \textbf{Product Images} & \textbf{Product Description}\\
    \midrule
    \pbox{3cm}{\centering \textit{Denim Parka with Genuine Fox Fur Trim}} & \pbox{7.5cm}{
    \raisebox{-.9\height}{\includegraphics[trim=0 5 0 0, width=40pt]{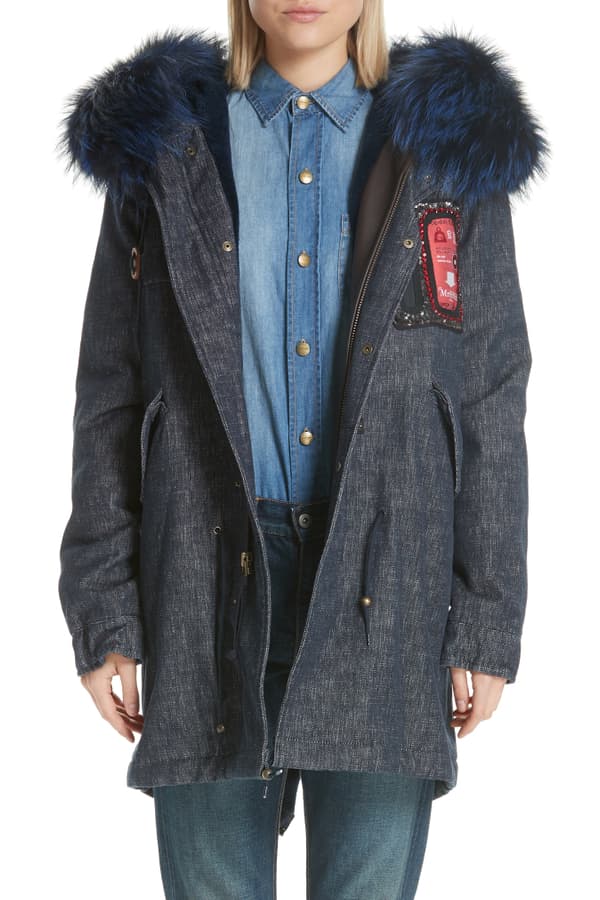}} \raisebox{-.9\height}{\includegraphics[trim=0 5 0 0, width=40pt]{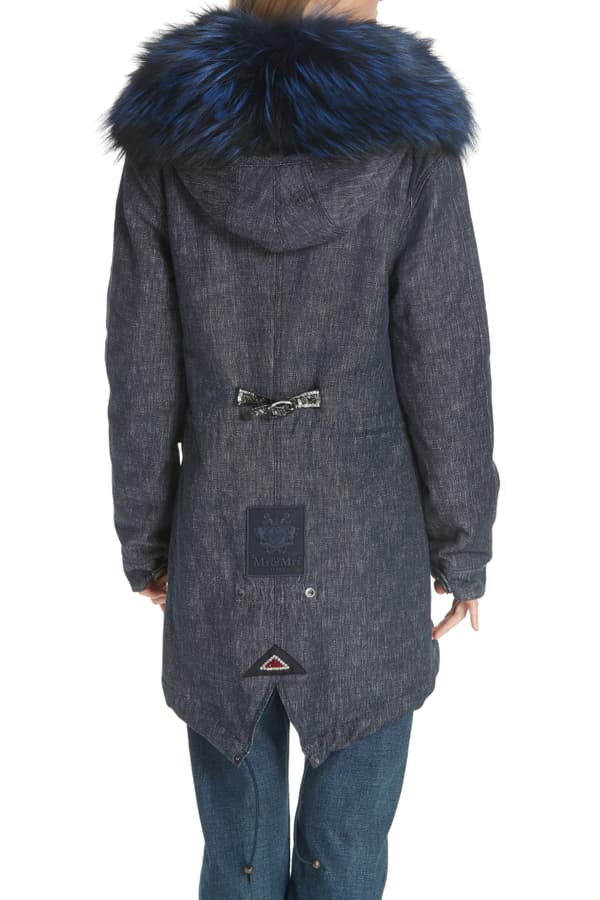}} \raisebox{-.9\height}{\includegraphics[trim=0 5 0 0, width=40pt]{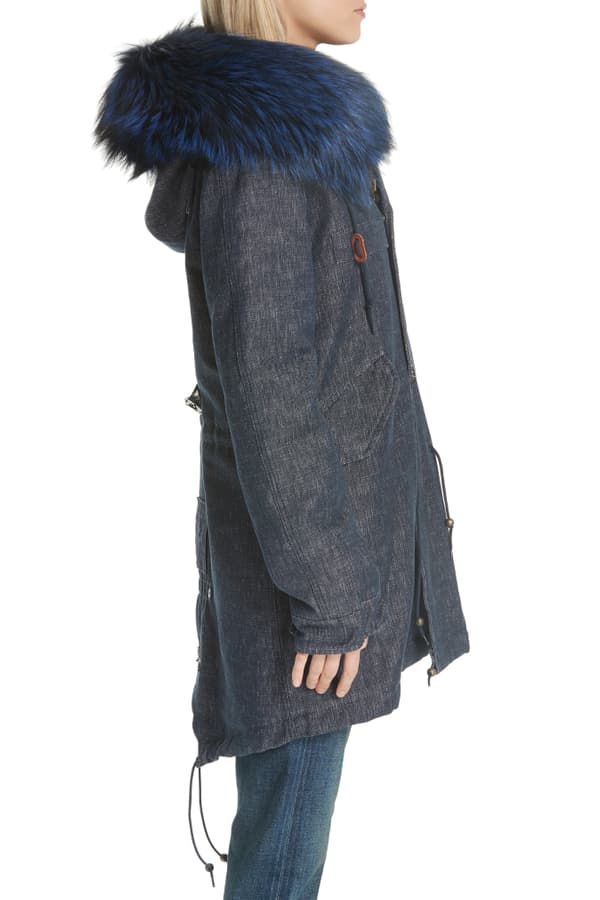}}
    \raisebox{-.9\height}{\includegraphics[trim=0 5 0 0, width=40pt]{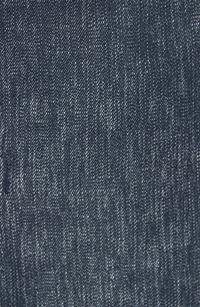}}
    \raisebox{-.9\height}{\includegraphics[trim=0 5 0 0, width=40pt]{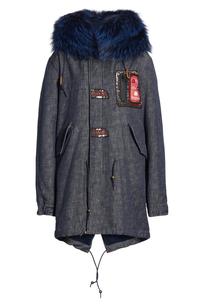}}
    } & \pbox{5cm}{\textit{Plush fox fur lining the hood and sparkling embellishments across the front bring luxe detail to a utilitarian-chic parka cut from pure-cotton Italian denim.}}\\
    \bottomrule
    \end{tabular}
\end{table*} 

\section{Related Work}
\label{related work}

\subsection{Transfer Learning in Multimodal Models}

Language representation model BERT \cite{devlin2018bert} demonstrated that transformer models trained with masked language modeling and next sentence prediction objective can lead to state-of-the-art performance for a variety of NLU tasks. VilBERT \cite{lu2019vilbert} extended the approach towards multimodality with separate transformer streams for image and text with cross-modality interaction though co-attention between the two streams. Other methods \cite{li2020unicoder, su2019vl, chen2020uniter, li2019visualbert} showed that single stream transformer models can learn the relationship between image and text. These models are pretrained on vision and language data, however \citet{kiela2019supervised} proposed a different approach where a pretrained unimodal (text) BERT model is fine-tuned together with a different modality (image), skipping the multimodal pretraining step. These methods are effective for understanding tasks like classification, but have not been studied for multimodal conditional generative tasks.

\subsection{Finetuning Natural Language Models for Controllability}

Unconditional language models can be adopted for text generation tasks such as language translation \cite{edunov2019pre}, question answering \cite{su2019generalizing}, and summarization \cite{zhang2019pretraining}. Other work demonstrates the sufficiency of providing the context text as the prefix in guiding text generation for these tasks \cite{brown2020language, radford2019language}. For example, translation models generate translated sentences given the source language input presented as prefix. Moreover, \citet{golovanov2019large} showed that concatenation of multiple textual contexts may form the guiding prefix. \citet{keskar2019ctrl} added controllability in language model during training by appending training corpus with different control codes, resulting in impressive generations for existing codes. However, these approaches are limited to text-based controllability.

\subsection{Conditionalizing Pretrained Language Models for Generation}


Models conditioned with pretrained language include noisy channel modeling \cite{yee2019simple} and fusion approaches \cite{sriram2017cold, gulcehre2015using} that concatenate hidden states of the conditional model with that of language model to predict the next word. Recently, \citet{ziegler2019encoder} proposed a modality invariant conditionalization approach for any transformer-based language model through pseudo self-attention. There, in every pretrained transformer layer the encoding vectors are considered as history and allowed to be attended over, leading to conditioning during self-attention. They also tested context attention where the decoder transformer layer is converted to a encoder-decoder, using pretrained weights for the decoding part. They demonstrated that pseudo self-attention is effective for even non-textual conditioning like image-based paragraph generation and class-based review generation. However, their work considers single modality conditioning whereas we treat the problem of multimodality conditioning. 

In this work, we use the approaches of \citet{ziegler2019encoder} as a comparative baseline. However, introducing new parameters within the whole network may hinder generative capabilities of pretrained language models. In addition, they require cumbersome manipulation of pretrained model architectures. MAnTiS addresses these issues with a simple approach requiring fewer additional parameters.\\
\begin{table*}
  \caption{Comparison of generator performance scores}
  \label{table2}
  \vskip 0.15in
  \centering
  \begin{tabular}{p{5cm}|cccc}
    \toprule
    \textbf{Model}     & \textbf{BLEU4}   & \textbf{CIDEr}     & \textbf{METEOR}     & \textbf{ROUGE-L}  \\
    \midrule
    \textsc{Context-Attn} & 3.9 & 30.4   & 10.2 & 17.2   \\
    \textsc{Pseudo-Self}     & 4.2  & 32.3  & 10.3 & 17.3 \\
    \textsc{MAnTiS} & \textbf{4.8} & \textbf{36.8} & \textbf{10.8} & \textbf{17.9} \\
    \midrule
    \textsc{MAnTiS-scratch} & 3.9 & 30.8 & 10.1 & 17.3 \\
    \textsc{MAnTiS-multi}  & 4.9 & 39.0 & 11.1 & 18.2 \\
    \textsc{MAnTiS-multi} + \textsc{text dropout}   & \textbf{5.0} & \textbf{39.5}  & \textbf{11.1} &
    \textbf{18.2} \\
    \bottomrule
  \end{tabular}
\end{table*}

\begin{table*}
  \caption{Qualitative evaluation of generated descriptions}
  \label{table3}
  \vskip 0.15in
  \centering
  \begin{tabular}{p{3cm}|ccccc}
    \toprule
    \textbf{Model}     & \textbf{Grammar}   & \textbf{Non-Redundancy}     & \textbf{Consistency}     & \textbf{Attractiveness} & \textbf{Overall}  \\
    \midrule
    \textsc{Context-Attn} & 0.74 & 0.89 & 0.75 & 0.51 & 0.603 \\
    \textsc{Pseudo-Self} & 0.76 & 0.93 & 0.72 & 0.54 & 0.578 \\
    \textsc{MAnTiS} & \textbf{0.82} & \textbf{0.96} & \textbf{0.81} & \textbf{0.62} &  \textbf{0.665} \\
    \bottomrule
  \end{tabular}
\end{table*}

\section{Method}
\label{method}
Given a sequence of token vectors $ x=(x_1, \dots, x_n) $, language models learn the probability $p(x)$, 
\[p(x) = \prod_{i=1}^n p(x_i \mid x_1, \dots, x_{i-1})\]
Here, we adapt a pretrained language model into a multimodal conditional model that learns the conditional probability distribution $p(x|y)$, where $y=(y_1, \dots, y_n)$ consists of tokens of any modality.
\[p(x|y) = \prod_{i=1}^n p(x_i \mid y, x_1, \dots, x_{i-1})\]
The goal is to learn $p(x|y)$ given supervised dataset of $x,y$ pairs. To achieve this, we frame the problem using an encoder-decoder architecture. We encode conditional modalities  using modality specific encoders and then project to the textual token space of the language model. Between the different modality types we add separator tokens, allowing the model to distinguish between them. We prepend these conditional tokens $y$ to the input guiding the generation $x$. We illustrate the overall architecture of our approach in Figure~\ref{fig1}.

The following subsections describe the encoding strategy, details of input construction, and the fine-tuning procedure.

\subsection{Encoder Mapping}
During the encoding stage we use both image and text modalities to condition the generation. To encode images we extract the embedding form of the last fully connected layer of a pretrained ResNet-152 model \cite{he2016deep}. This can be regarded as a single dense token per image whose dimension $N$ depends on the ResNet model. Transformation of the input image uses the same setting as during the pretraining process, which includes resizing, center cropping and normalization. Next, we project the token into the language model embedding space $D$ through a linear layer with learnable weight matrix $W \in \mathcal{R}^{N \times D}$.\\

The embedding function of the decoder language model encodes the text. For the language model, we use the transformer-based pretrained model GPT-2 \cite{radford2019language}, an auto-regressive model whose self-attention module can attend only on previous tokens, with Byte Pair Encoding (BPE) for text  tokenization.


This approach can easily be extended towards any modality because we map the encoding to the textual space. The encoder and decoder are jointly fine-tuned end-to-end during supervised learning. Allowing the encoder, specifically, the image encoder to be fine-tuned will contribute towards effective learning of image token mapping.

\subsection{Multimodal Fine-tuning}

In the GPT-2 language model the input consists of a sum of token and position embeddings, with the position encoding zero-indexed. For each conditional modality, we start the position encoding from zero as well. Between each modality token we add a separator token [SEP] whose position is one plus the previous token position. The first conditional token is prepended with a beginning of sentence [BOS] token and the generation ends with a end of sentence [EOS] token.

During fine-tuning, the model is trained using the same loss function (cross-entropy) as GPT-2, between the next predicted word of the language model head and the ground truth word. No loss is computed for the image tokens because they have no exact vocabulary.

\subsection{Modality Dropout}

Fine-tuning in this manner forces the pretrained language model to learn cross-modality correlations between image and text. Naturally, this can cause text tokens to influence generation more than other modality tokens. In our approach, we pretrain the decoder language model and image embedding model while we randomly initialize the image mapping layer for training. \citet{neverova2015moddrop} proposed ModDrop, arguing that randomly dropping different modality channels during training could help learn cross-modality representations and reduce false co-adaptions.

Because image representations must be fused into the text-only model, we randomly dropped out text conditionality paths with a probability $p$ tuned during training. We speculate that this could provide improved image conditioning and lead to better overall performance. This was performed in addition to the standard dropout within the transformer decoder layers.

\begin{table*}
  \caption{Sample outputs on the FACAD dataset.}
  \label{table4}
  \vskip 0.15in
  \centering
{\renewcommand{\arraystretch}{1.25}%



\begin{tabular}{p{3cm}|p{2.5cm}|p{7.cm}}
\toprule
    \textbf{Input} & \textbf{Model} & \textbf{Generated Text}\\
    \midrule
\multirow{3}{*}{
\pbox{3cm}{\centering Beverly Skinny Flare\phantom{-}Jeans
\includegraphics[trim=0 0 0 -10, width=56pt]{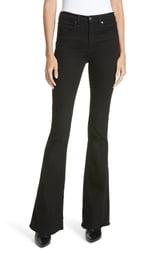}}
} &  \textsc{Context-Attn} & \textit{Tonal stitching and subtle fading add worn in character to dark wash jeans cut with a flattering figure flattering flare.} \\\cline{2-3}
    & \textsc{Pseudo-Self} & \textit{These stretchy bootcut jeans are inspired by the '70s super-sweet denim look that has you feeling like the real thing.}\\\cline{2-3}
    & \textsc{MAnTiS} & \textit{Figure-flattering flared jeans are made from soft denim with shape-retaining stretch and a clean front for a modern silhouette.}\\\midrule 
\multirow{3}{*}{
\pbox{3cm}{\centering Azur Tassel Hem Cotton Blouse 
\includegraphics[trim=0 0 0 -10, width=58pt]{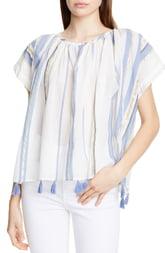}}
} & \textsc{Context-Attn} & \textit{With a twirl that's T-shirt cut to the natural waist, this gauzy blouse is ready to be fancy or just fun.} \\\cline{2-3}
    & \textsc{Pseudo-Self} & \textit{From a collaboration with fashion/lifestyle blogger Lindsey Schuster, cute pieces like this top prove that looking good can be a breeze—even on crazy-busy days.} \\\cline{2-3}
    & \textsc{MAnTiS} & \textit{Shine through your work-to-play look in this gauzy blouse trimmed with embroidered tassels for a free-spirited vibe.}\\
\bottomrule
\end{tabular}

}
\end{table*}

\section{Experimental Setting}
\label{experimental setting}
This section includes detailed information on the datasets, metrics, and baselines used during training and evaluation.

\subsection{Datasets}

We used the initially released version of the fashion captioning dataset FACAD \cite{yang2020fashion}. This dataset consists of fashion articles and their names, images from different perspectives, e-commerce relevant descriptions, colors, and other pieces of metadata. In this work, we want to generate product descriptions given the title and various images of a product. An example of the dataset is shown in Table~\ref{table1}. There are total 55,959 descriptions. We removed entries with empty description, name or images, as well as duplicated descriptions, reducing the size to 45,748. Out of these 40,748 were used for training, 2,500 for validation, and the remaining 2,500 for testing.

\citet{yang2020fashion} used this dataset for the image captioning problem where the generated caption depends only on a single given image. We used this dataset for multimodal conditioned NLG, where multiple instances of each modality may be provided as input. However, using multiple images per description significantly reduces the total number of training samples.

\subsection{Evaluation Metrics}

To perform qualitative evaluation we report model performance on the most commonly used NLG metrics, which include BLEU4 \cite{papineni2002bleu}, CIDEr \cite{vedantam2015cider}, METEOR \cite{denkowski2014meteor}, and ROUGE-L \cite{lin2004rouge} scores.

\subsection{Training Details}

GPT-2 is a large transformer-based model trained on the WebText dataset (${\sim}40$ GB), consisting of text from 8 million non-Wikipedia webpages \cite{radford2019language}. It shows excellent performance with coherent text generation; thus, we use it as the base unconditional pretrained language model. In particular, we use GPT-2 medium, possessing an embedding size of $1024$, comprising 24 layers with 16 heads per layer and including a total of 345M parameters. It is publicly available from the HuggingFace repository \cite{wolf2019huggingface}. We use the same vocabulary with an addition of three tokens: BOS, SEP and PAD (padding token). For encoding images we use ResNet-152 trained on the ImageNet dataset \cite{deng2009imagenet}, which is publicly available in PyTorch's torchvision package \cite{paszke2019pytorch}.

Additionally, we tuned the learning rates for each model between $1\mathrm{e}{-5}$
 to $5\mathrm{e}{-5}$. We tuned text modality dropout between $0.3$ to $0.7$ and set all other dropout values to 0.1. Training was done using the AdamW optimizer and a linear scheduler with warmup.

\subsection{Baseline Methods}

We compare MAnTiS against the current most advanced approaches for conditioning language model. In comparing approaches we used the latest available code published by the authors.


\textsc{Context-Attn}: Context attention adds a randomly initialized encoder-decoder layer on top of every pretrained decoder layer of GPT-2 \cite{ziegler2019encoder}. Multimodal conditionality tokens are used as the encoder tokens.

\textsc{Pseudo-Self}: Pseudo self-attention prepends additional multimodality conditioning tokens to every self-attention layer of GPT-2 \cite{ziegler2019encoder}. This achieved the best performance for uni-modal conditioning, forming the strongest baseline.

\begin{table*}
  \caption{MAnTiS generated descriptions with colored annotations (green: higher quality; blue: information unique to the image; purple: textual information; orange: incorrect information).}
  \label{table5}
  \vskip 0.15in
  \centering
  \begin{tabular}{p{2.5cm}p{4.5cm}|p{5.5cm}}
    \toprule
    \textbf{Input} & \textbf{} & \textbf{Generated Text}\\
    \midrule
    \pbox{2.5cm}{\centering \textit{Cover-Up Dress}} & \pbox{4.5cm}{
    \raisebox{-.9\height}{\includegraphics[trim=0 5 0 0, width=40pt]{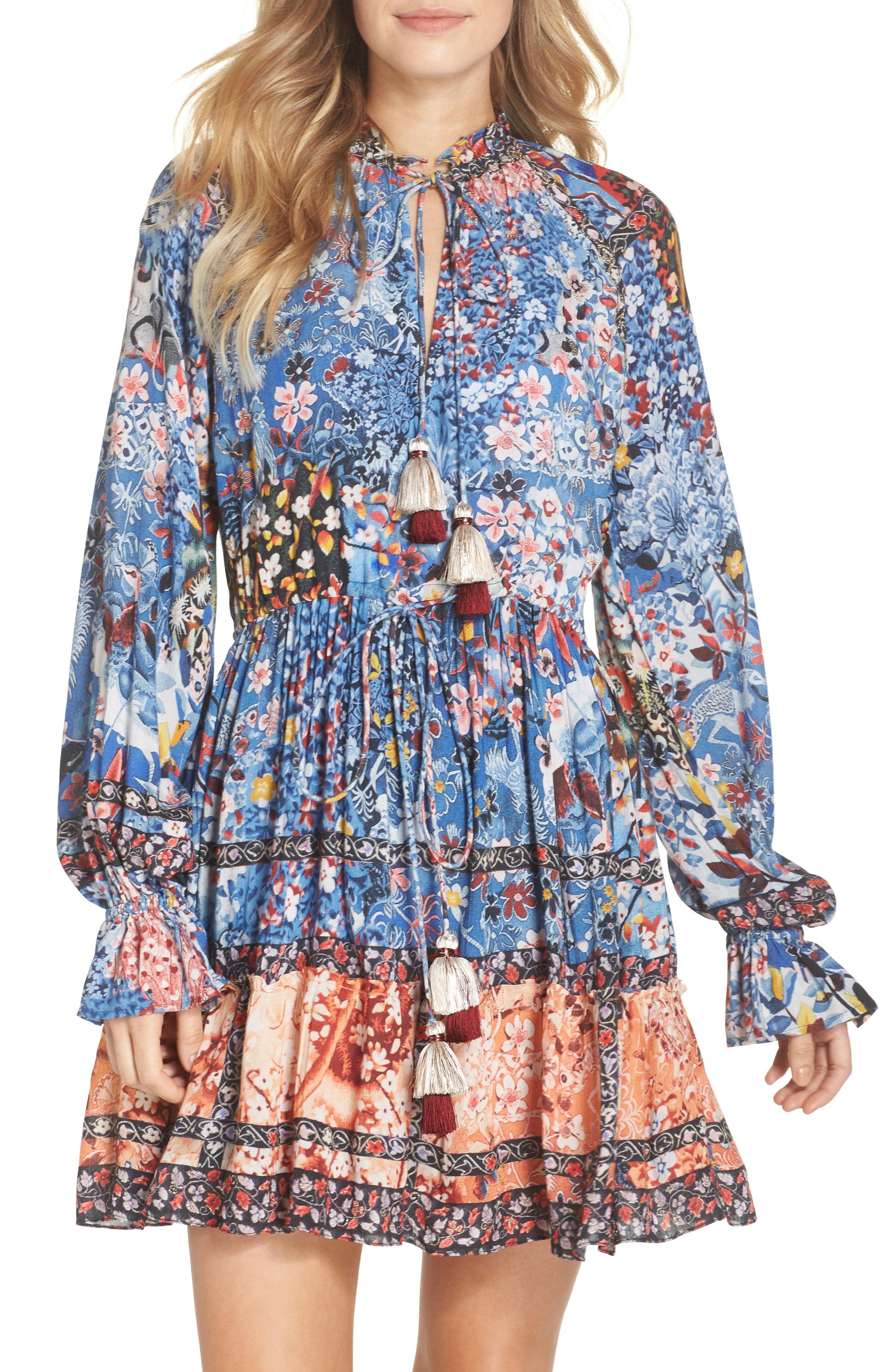}} \raisebox{-.9\height}{\includegraphics[trim=0 5 0 0, width=40pt]{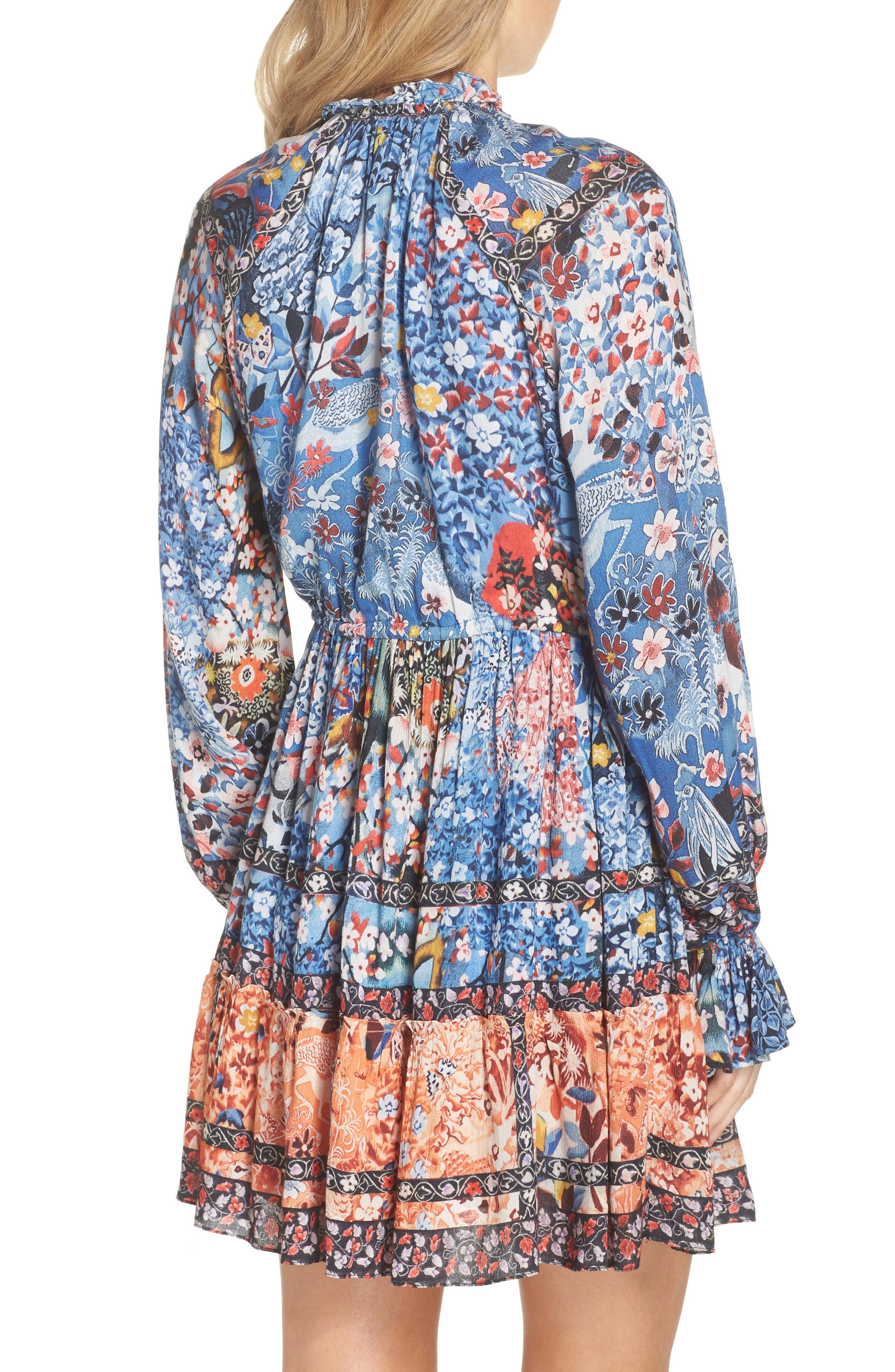}} \raisebox{-.9\height}{\includegraphics[trim=0 5 0 0, width=40pt]{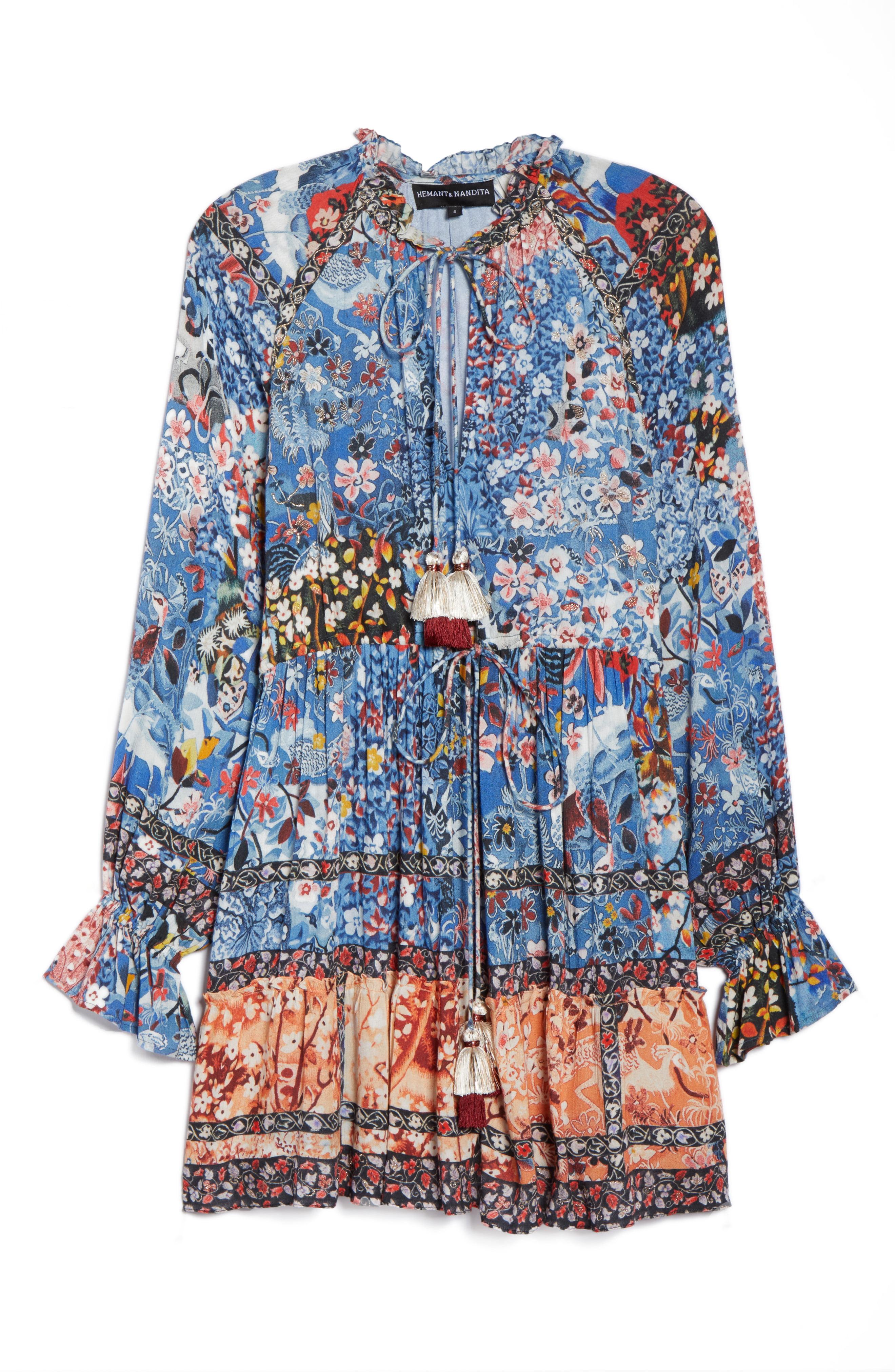}}} & \pbox{5.5cm}{\textit{If you don't already have \textcolor{ForestGreen}{a beach vacation planned}, this \textcolor{NavyBlue}{long-sleeve} \textcolor{Plum}{cover-up dress} in a vintage \textcolor{NavyBlue}{floral print} would like you to reconsider that.}}\\
    \midrule
    \pbox{2.5cm}{\centering \textit{Velour-Hooded Jumpsuit}} & \pbox{4.5cm}{
    \raisebox{-.5\height}{\includegraphics[trim=0 0 0 0, width=40pt]{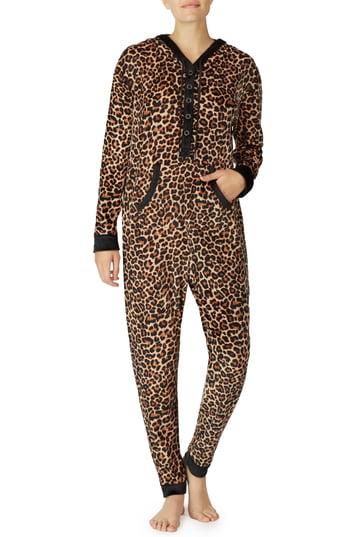}} \raisebox{-.5\height}{\includegraphics[trim=0 0 0 0, width=40pt]{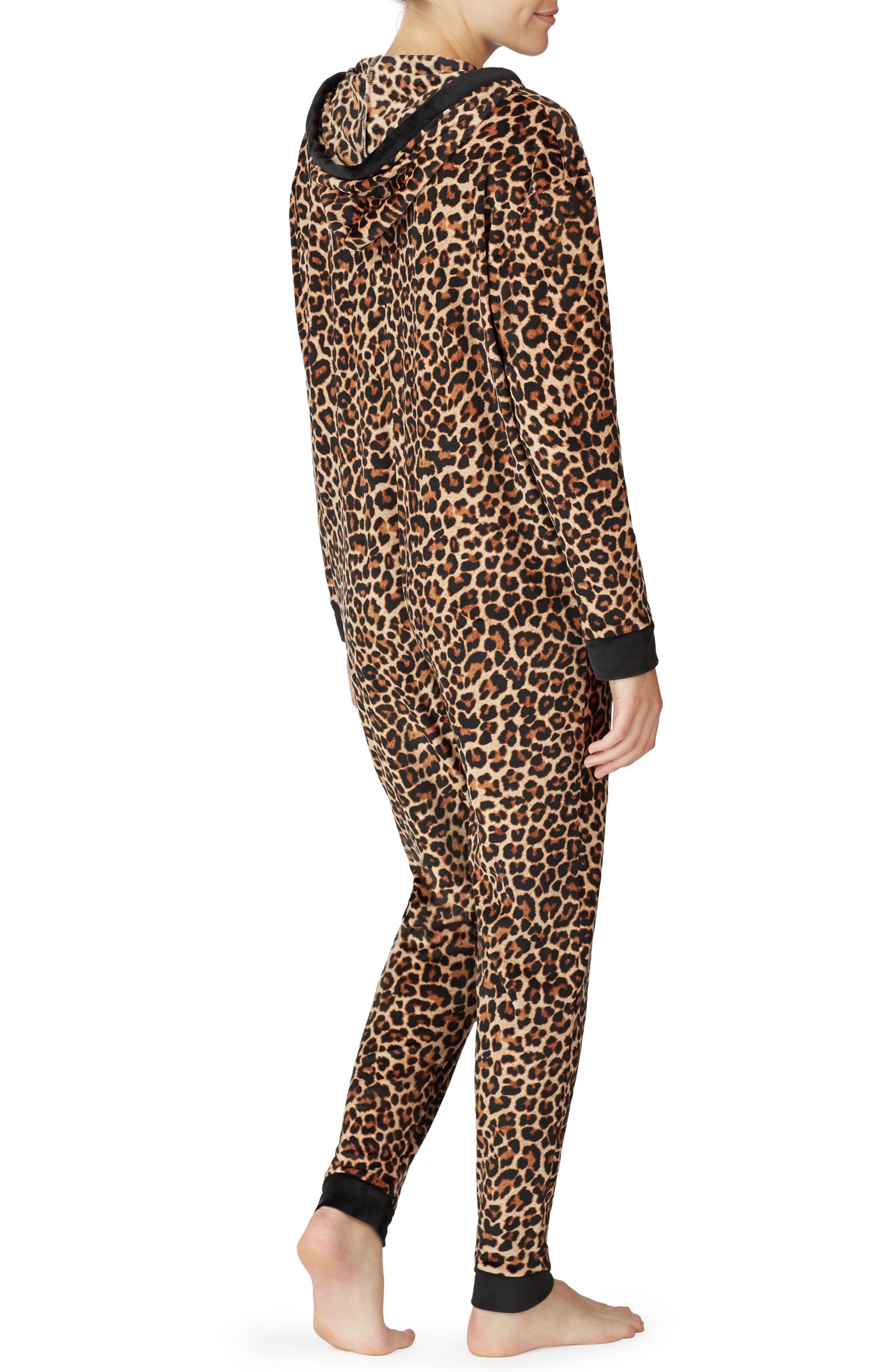}} \raisebox{-.5\height}{\includegraphics[trim=0 0 0 0, width=40pt]{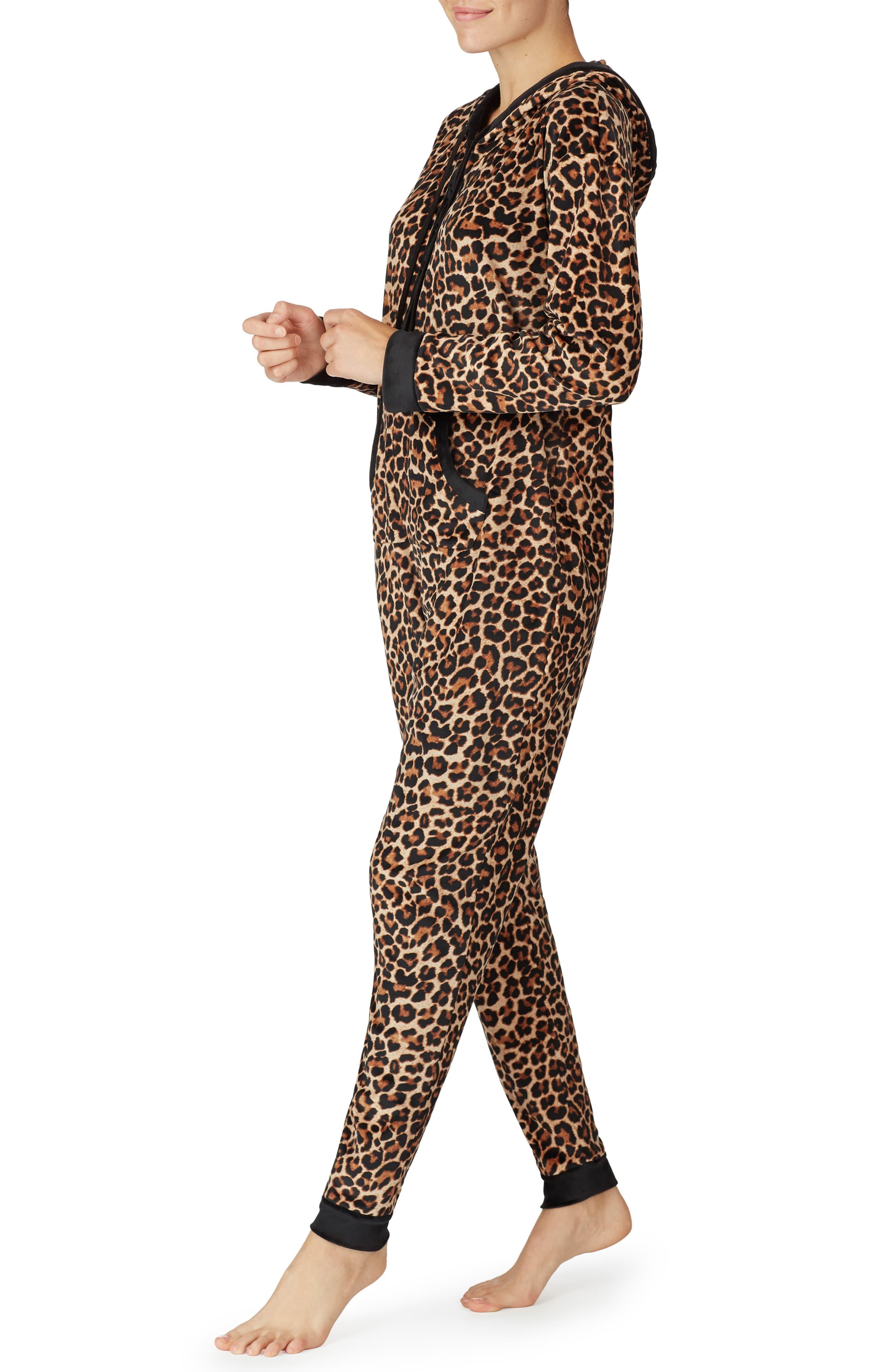}}} & \pbox{5.5cm}{\textit{Get a \textcolor{ForestGreen}{luxe look in this one-and-done} \textcolor{Plum}{jumpsuit} designed in \textcolor{ForestGreen}{sumptuous velour} with a dramatic high/low hem.}}\\
    \midrule
    \pbox{2.5cm}{\centering \textit{Hudson Holly High Waist Distressed Deconstructed Crop Flare Jeans}} & \pbox{4.5cm}{
    \raisebox{-.5\height}{\includegraphics[trim=0 0 0 0, width=40pt]{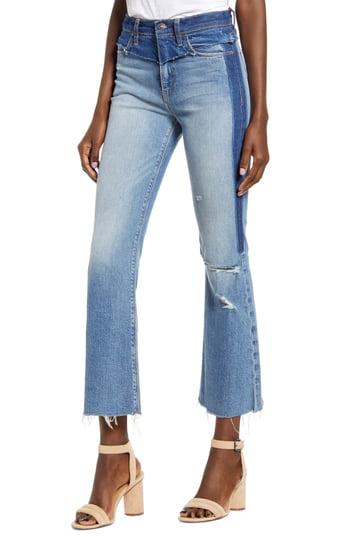}} \raisebox{-.5\height}{\includegraphics[trim=0 0 0 0, width=40pt]{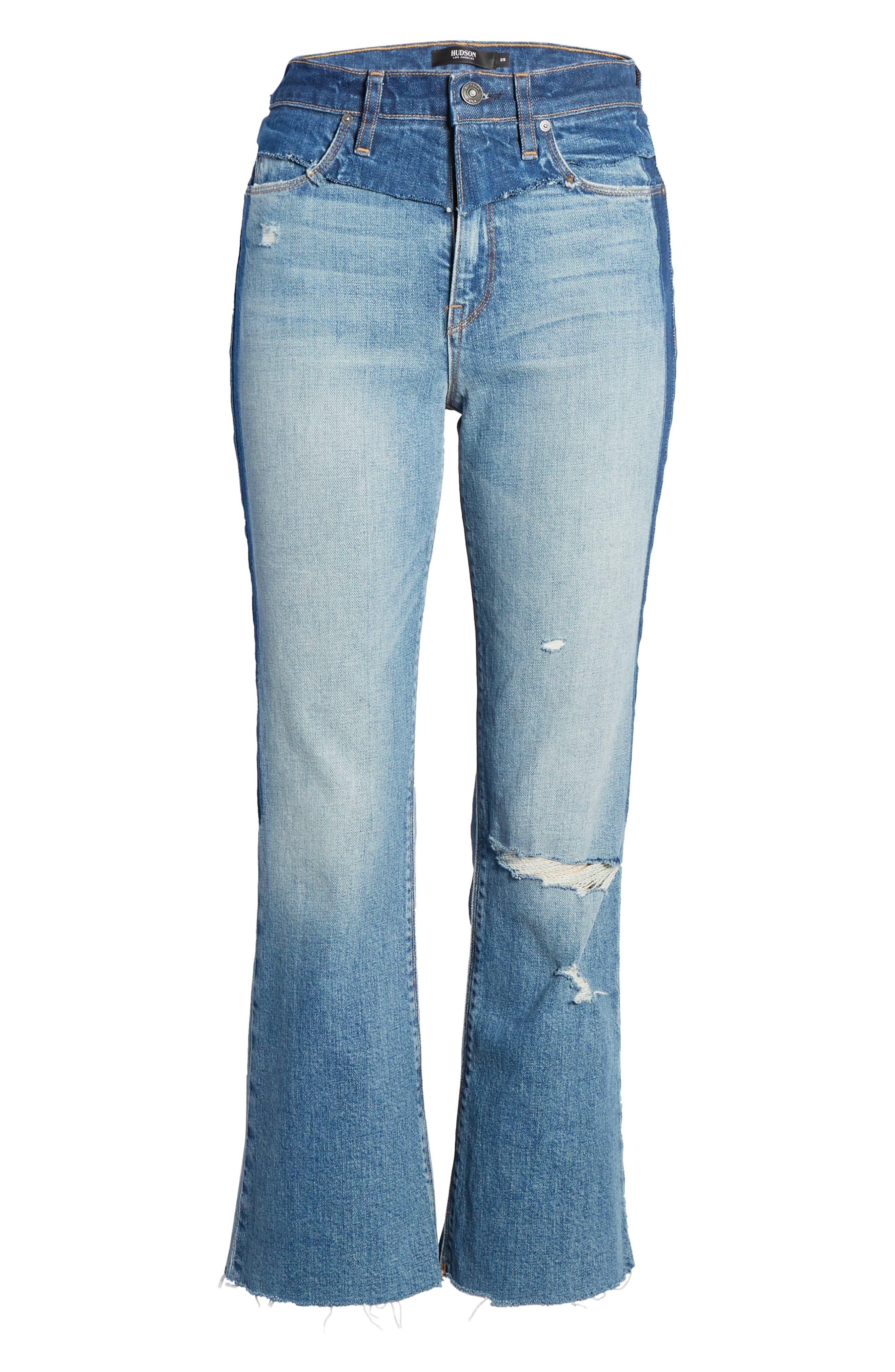}} \raisebox{-.5\height}{\includegraphics[trim=0 0 0 0, width=40pt]{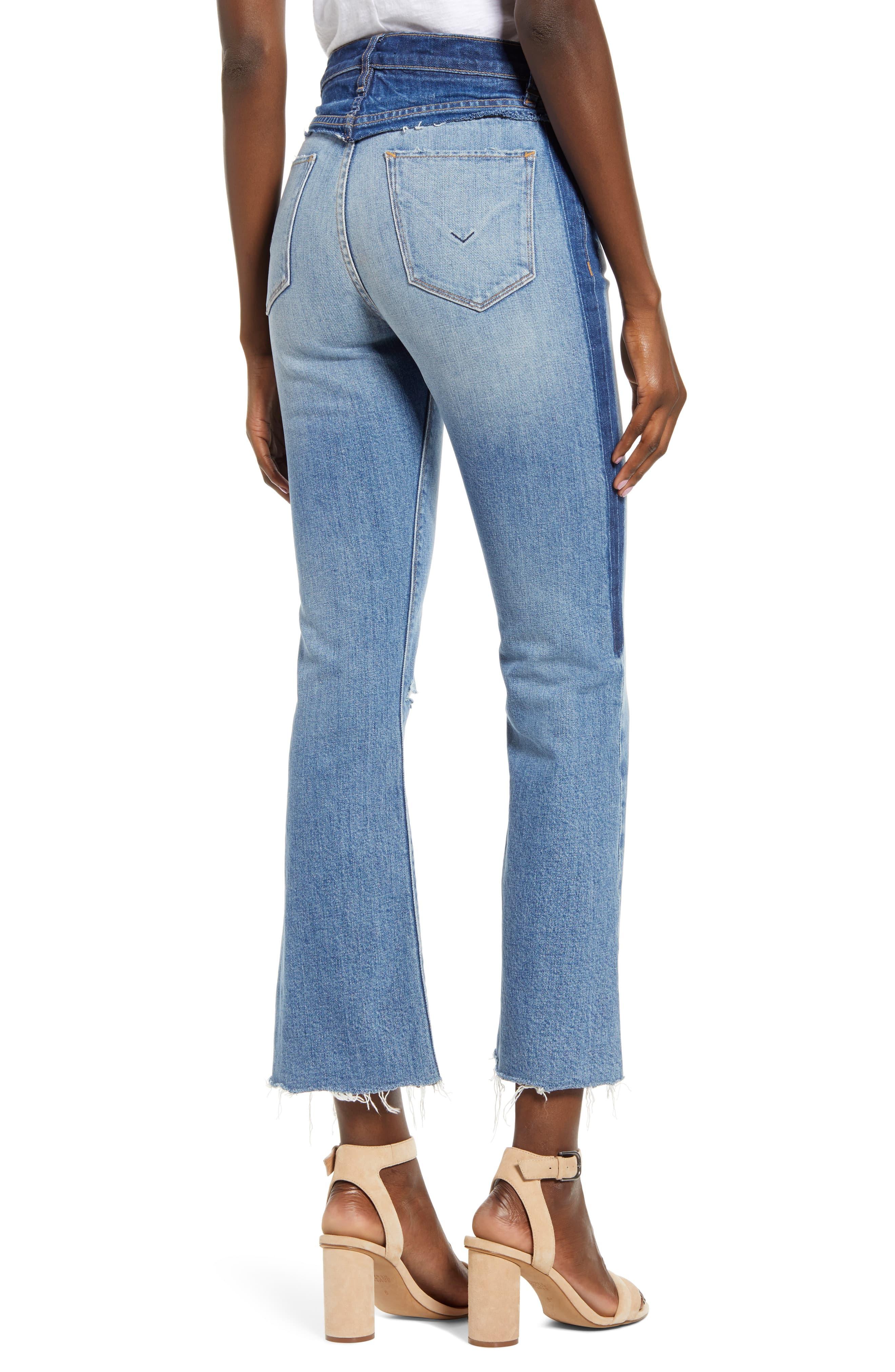}}} & \pbox{5.5cm}{\textit{Essential \textcolor{Orange}{white jeans} get dashed with destruction, from the \textcolor{NavyBlue}{ripped knee to the slashed hem}, and the end result \textcolor{ForestGreen}{delivers some drama} for denim days and nights.}}\\
    \midrule
    \pbox{2.5cm}{\centering \textit{Davis Feather Trim Cami}} & \pbox{4.5cm}{
    \raisebox{-.5\height}{\includegraphics[trim=0 0 0 0, width=40pt]{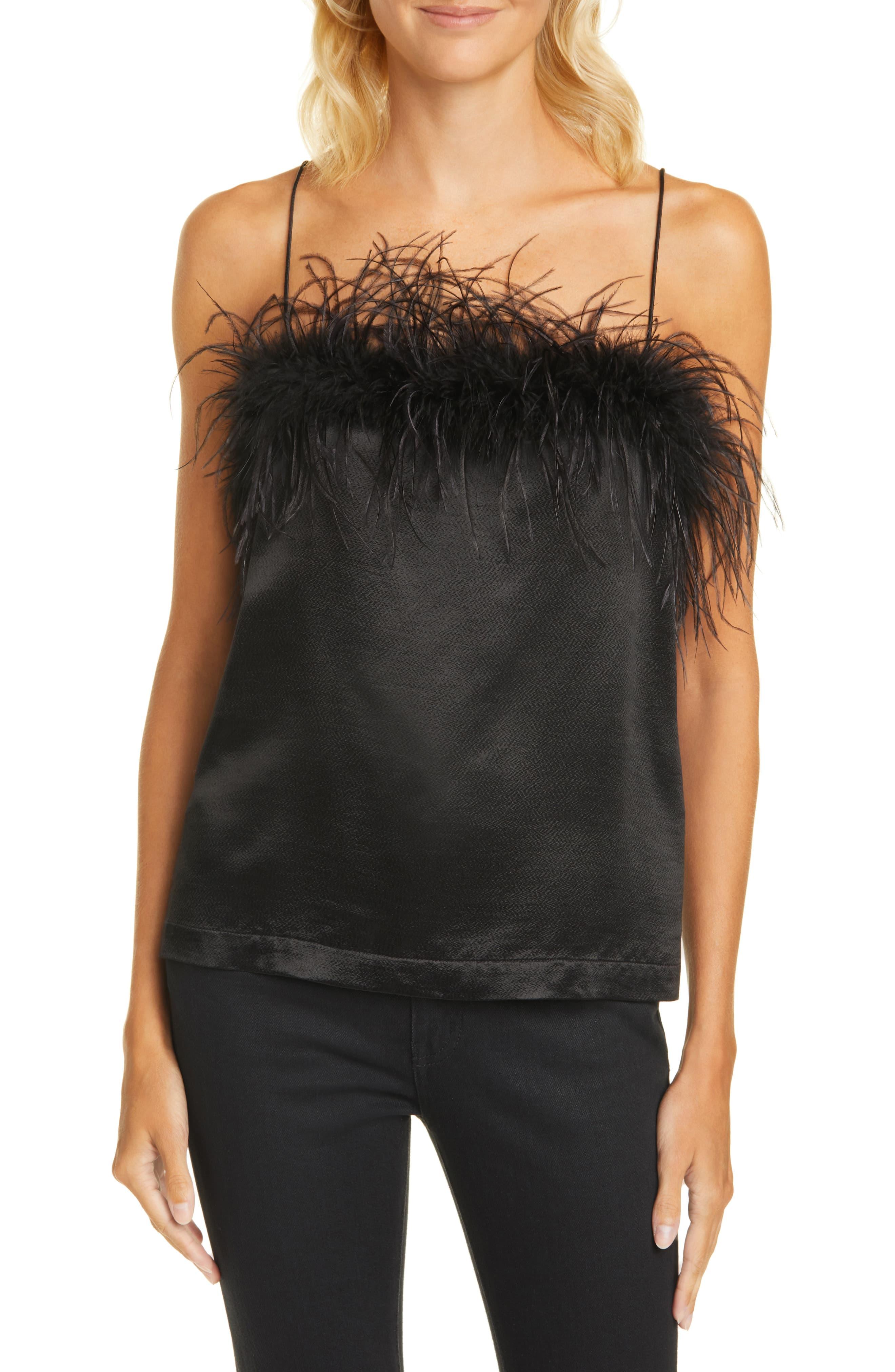}} \raisebox{-.5\height}{\includegraphics[trim=0 0 0 0, width=40pt]{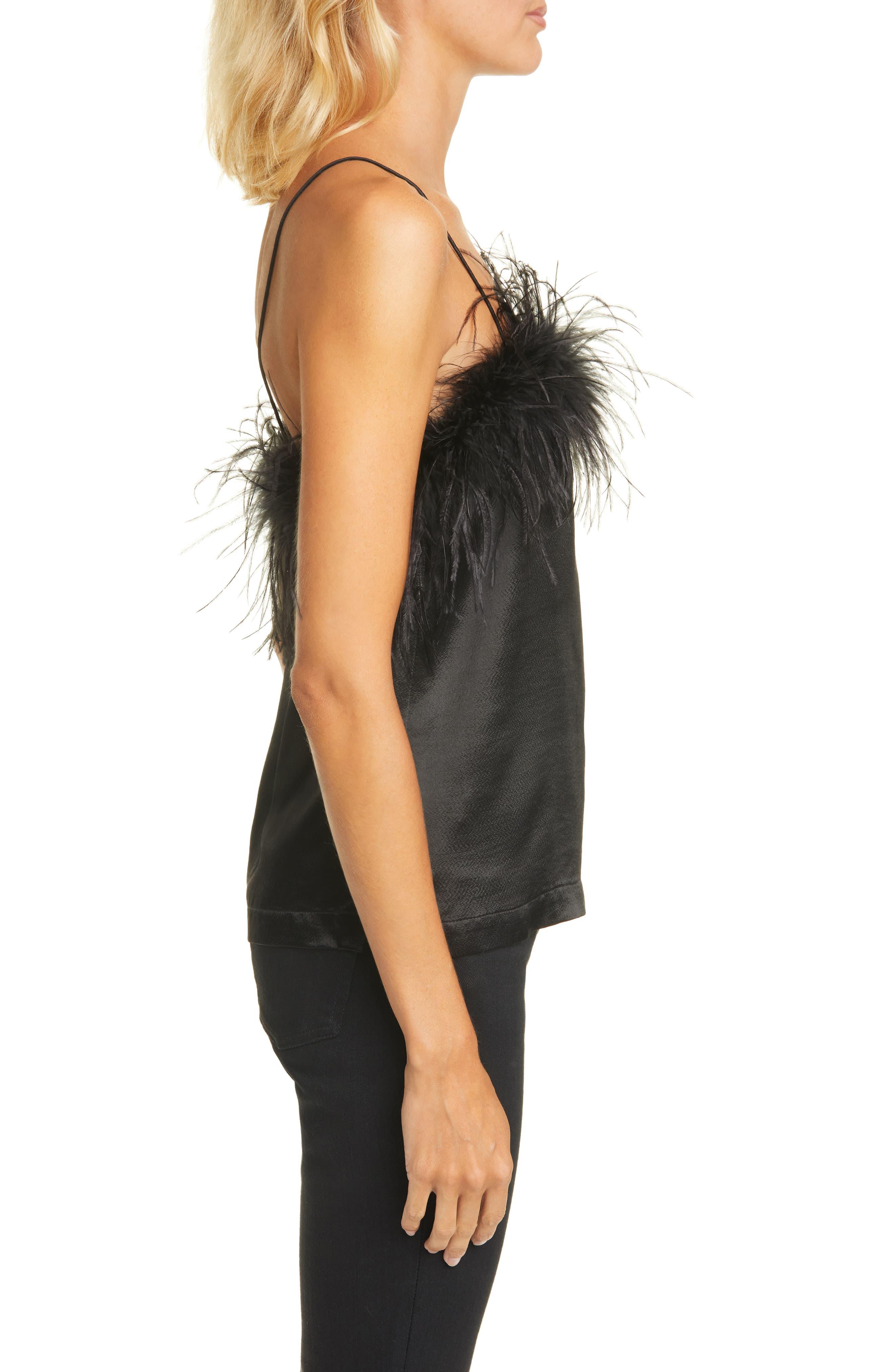}} \raisebox{-.5\height}{\includegraphics[trim=0 0 0 0, width=40pt]{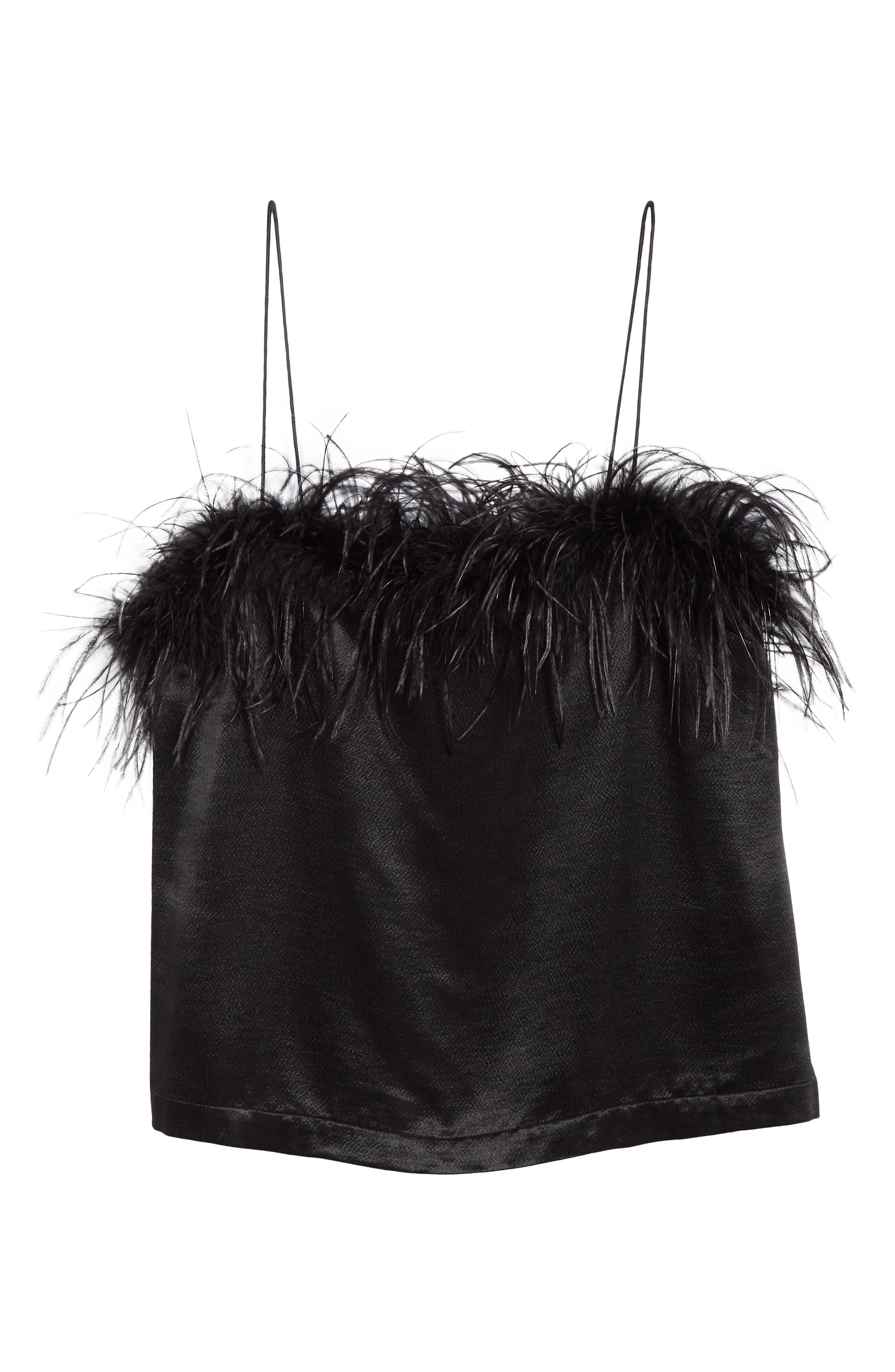}}} & \pbox{5.5cm}{\textit{A \textcolor{Plum}{feather-trimmed} hem and \textcolor{NavyBlue}{spaghetti straps} add \textcolor{ForestGreen}{dynamic finishing touches} to this streamlined \textcolor{Plum}{cami}.}}\\
    
    \bottomrule
    \end{tabular}
\end{table*}  

\section{Results}
\label{result}

A single image for each fashion item was chosen (the first image in the dataset sample) to fairly compare Context-Attn, Pseudo-Self, MAnTiS, and MAnTiS-scratch. In MAnTiS-scratch we randomly initialized the internal language model. In MAnTiS-multi we used up to five images if available and added conditioning text dropout with a probability $0.3$. We used the full product name of each fashion item to train all models.


\subsection{Fashion Description Generations}

The main quantitative results for the fashion description generation task are summarized in Table~\ref{table2}. MAnTiS significantly outperformed the baseline approaches in all the evaluation metrics. MAnTiS improved the BLUE4 score by 0.6, CIDEr by 4.5, METEOR by 0.5, and ROUGE-L by 0.6 respectively compared to Pseudo-Self, which shows the effectiveness of our approach.



Context-Attn adds a new layer in each transformer block, and jointly optimizing pretrained weights along with the newly initialized weights in every layer adds difficulty. We suspect this hindered the information gain from images, or even negatively impacted the generation capability of the pretrained model. Similar observations were made in uni-modal settings by \citet{ziegler2019encoder}. We believe our approach outperformed the strong baselines because it entailed minimal interference with the pretrained language model. We introduced new parameters only at the input level, compared to Pseudo-Self which alters the self-attention module.

We next analyzed whether using multiple images improved text generation. The dataset provides several images of each fashion article. Providing MAnTiS-multi with up to 5 images increased performance over MAnTiS in all metrics, with improvements of 0.1, 2.2, 0.3 and 0.3 in BLEU4, CIDEr, METEOR and ROUGE-L respectively. This shows that the model combines information from different visual inputs. We further studied the effect of incorporating text modality dropout. Dropping out product name information randomly improved the performance slightly over MAnTiS-multi in BLEU4 and CIDEr, with no change in METEOR or ROUGE-L. This indicates that modality dropout can provide small benefits and no negative effect during optimization. As expected, MAnTiS-scratch greatly underperformed MAnTiS-single, indicating the benefit of large pretrained models.


\subsection{Human Rating}

Product descriptions geared towards e-commerce should entice customers using appealing phrases. Difficulty arises when analyzing such properties using automatic metrics, so we performed human evaluations to rate aspects of generated descriptions. Two random judges of different genders were tasked to score 200 product descriptions. Inspired by \citet{dang2005overview} we asked the raters to measure five linguistic qualities including grammar, non-redundancy, consistency, attractiveness, and overall scores. For the first four qualities, the task demanded only a yes/no answer. A consistent description is coherent and correct given a product's name and image, and is attractive if it is interesting or attention-grabbing. For the last category ``overall'', judges scored descriptions between 1 (worst) and 5 (best) from an e-commerce perspective.

The normalized average scores are shown in Table~\ref{table3}. MAnTiS outperformed the baseline approaches in all five categories. This shows that our conditional adaptation approach is significantly better than the previous approaches. This is likely due to the fact that MAnTiS does not introduce new parameters within the pretrained language model, unlike other approaches. 


\subsection{Qualitative Analysis}

In Table~\ref{table4} we show example generations from different models. We see that Context-Attn has repetition and incorrect information like ``subtle fading'', while no fading is seen in the image. MAnTiS generated higher quality descriptions highlighting image features like ``trimmed with embroidered tassels''.


To illustrate our results, we give a few representative MAnTiS generations from the test dataset in Table~\ref{table5}. We color-coded important parts of the generated text with green to indicate a high-quality phrase, blue to indicate attributes present only in the image, purple to indicate attributes from product name and orange to indicate possibly incoherent information. The blue highlighted phrases demonstrate that MAnTiS generations are guided in part by image content. In the first row, the generated description aptly connects the Cover-Up Dress to a beach setting. The model may sometimes fail to pick up on image-based cues correctly, as seen in third example where color was pronounced as ``white'' instead of denim blue, although this confusion is understandable as the faded regions are white. Overall, the examples show that MAnTiS can generate diverse coherent descriptions conditioned on both modalities.

\section{Conclusion}
\label{conclusion}

In this work, we introduce MAnTiS, a novel approach for adapting pretrained language models into multimodal conditional NLG models. We showed that our approach significantly outperforms strong baselines methods on several common NLG evaluation metrics. For a qualitative analysis, we perform human evaluations and show that our approach generates high-quality text that agrees with the conditional input. Based on several qualitative measures we show that conditionalizing a pretrained language model through new modalities does not hamper its generative capabilities.

Our approach is straightforward, easy to implement, and extendable to any modality and provides an effective way to conditionalize any pretrained language model. We believe this study will set a strong baseline in the field of multimodal NLG.

\nocite{langley00}

\bibliography{example_paper}
\bibliographystyle{icml2021}

\end{document}